# Assessing and Predicting Air Pollution in Asia: A Regional and Temporal Study (2018-2023)

Anika Rahman[1] and Dr. Mst. Taskia Khatun[2]

[1]Department of Computer Science and Engineering, Stamford University Bangladesh, Dhaka, Bangladesh
[2]Department of Software Engineering, Daffodil International University, Dhaka, Bangladesh

## ABSTRACT

*This study analyzes and predicts air pollution in Asia, focusing on PM 2.5 levels from 2018 to 2023 across five regions: Central, East, South, Southeast, and West Asia. South Asia emerged as the most polluted region, with Bangladesh, India, and Pakistan consistently having the highest PM 2.5 levels and death rates, especially in Nepal, Pakistan, and India. East Asia showed the lowest pollution levels. K-means clustering categorized countries into high, moderate, and low pollution groups. The ARIMA model effectively predicted 2023 PM 2.5 levels (MAE: 3.99, MSE: 33.80, RMSE: 5.81, R²: 0.86). The findings emphasize the need for targeted interventions to address severe pollution and health risks in South Asia.*

## 1. INTRODUCTION

Air pollution, particularly PM 2.5 (particulate matter with a diameter of less than 2.5 micrometers), is a significant global environmental and public health issue due to its severe health impacts, such as respiratory and cardiovascular diseases and premature death. Asia, with over 60% of the global population, faces some of the highest PM 2.5 levels, especially in South Asia, where several cities rank among the most polluted worldwide. In recent years, the rapid industrialization, urbanization, and population growth in Asia have exacerbated air quality issues, particularly in densely populated regions [1]. This study analyzes PM 2.5 trends across five Asian regions— Central, East, South, Southeast, and West Asia—from 2018 to 2023. Data from 38 countries were analyzed to explore regional pollution disparities, correlations between PM 2.5 levels, population density, and mortality rates, and classify countries into pollution categories using K-means clustering. The ARIMA model was employed to predict PM 2.5 levels, achieving robust metrics (MAE: 3.99, MSE: 33.80, RMSE: 5.81, R²: 0.86) for 2023 predictions. Insights from this study aim to support policy interventions and public health measures to mitigate air pollution impacts. The paper is organized as follows: Section 2 reviews relevant literature; Section 3 outlines the dataset and methodology; Section 4 presents results on PM 2.5 trends and correlations, details the predictive modelling; and Section 5 concludes with implications and future research directions.

## 2. LITERATURE REVIEW

Over recent decades, air pollution in Asia has become a major threat to food security [2] and human health [3]. Wildfire smoke, pollen-based aeroallergens, and climate change, primarily due to greenhouse gas emissions, heavily influence PM2.5 levels. Research highlights severe impacts on health and the environment, with 37 of the world's 40 most polluted cities in South Asia [4]. Contaminated air in this region leads to millions of preventable deaths annually and harms crops essential for feeding many [5]. Biomass burning is a significant source of haze [6], accounting for up to 40–60% of haze events in Southeast Asia from 2003 to 2014 [7]. This worsening air quality presents significant challenges for sustainability and health. In South Asia, countries like India, Nepal, Bangladesh, and Pakistan face severe declines in air quality due to climate change [8]. Additionally, air pollution contributes to about 349,681 pregnancy losses annually [9] and impacts food security, with estimates suggesting that it damages crops sufficient to feed 94 million people each year and reduces life expectancy by approximately three years for about 660 million people [10]. Several models have been created to forecast these potential consequences. However, making precise forecasts is very impossible. A hybrid intelligent model integrating LSTM and MVO has been created to forecast air pollution from Combined Cycle Power Plants [11]. A deep learning framework using a temporal sliding LSTM extended model has been created [12].

Recent studies have advanced air quality prediction models by integrating machine learning techniques. One model, VMD-CSA-CNN-LSTM, combined CNN and LSTM networks [13] for AQI prediction in nine Chinese cities. Optimized with the Chameleon Swarm Algorithm and variational mode decomposition, it achieved high accuracy (RMSE of 2.25, adjusted R-squared above 96%). PM2.5 and PM10 were identified as key pollutants, with ozone also significant in some cities, providing a tool for air quality management. Another study in Varanasi, India, compared machine learning models [14] like Random Forest, Decision Tree, and SVM to predict AQI based on PM2.5, PM10, CO, NO2, NH3, SO2, and O3 levels. Random Forest and Decision Tree models outperformed others, achieving near-perfect accuracy, highlighting their potential for real-time air quality prediction and urban planning.

## 3. METHODOLOGY

### 3.1. Dataset Description

This study uses air quality data from IQAir [15] covering PM 2.5 levels in Asia from 2018 to 2023. The dataset includes columns for Region, Country, annual PM 2.5 values, Population (2023), Area, and death rates from air pollution (2018-2021). Initially covering 38 countries, the dataset was refined to 36 countries for ARIMA model predictions due to data availability.

### 3.2. Data Preprocessing

1) Handling Non-Numeric Values: PM 2.5 values were converted to numeric, with non-numeric entries replaced by NaN and then filled with 0.
2) Population and Area Data Cleaning: Removed commas and converted Population and Area data to numeric format.
3) Population Density Calculation: A new feature, Population Density, was calculated by dividing the Population by the Area for each country.
4) Filtering: Only Asian countries were included, and missing values in PM 2.5 data were filled with zeros.
5) Regional Classification: Countries were categorized into one of the five Asian regions.

### 3.3. Temporal Analysis

To analyze the temporal trends of PM 2.5 levels across Asia from 2018 to 2023, we conducted the following steps:

1) Regional Analysis: We created individual time series plots for PM 2.5 levels across the five Asian regions (Central, East, South, Southeast, and West Asia) to observe regional patterns. Additionally, we conducted trend analysis by plotting PM 2.5 levels over time for each country within the regions, allowing us to compare temporal changes and identify regional trends.
2) Overall Asia Analysis: We calculated average PM 2.5 levels for each region and generated a summary plot to provide an aggregated view. This summary plot enabled us to compare temporal trends across all five regions, offering a comprehensive overview of air pollution evolution in Asia from 2018 to 2023.

### 3.4. Death Rate Analysis

To examine the impact of air pollution on mortality rates across different regions in Asia, we conducted a detailed analysis of death rates attributable to air pollution from 2018 to 2021.

1) Regional Death Rate Analysis: We calculated and visualized average death rates per region for each year (2018-2021) using bar plots. This approach allowed for a clear comparison of mortality rates across regions and highlighted disparities in death rates.
2) South Asia Mortality Trends: We focused on South Asia due to its high PM 2.5 levels and death rates. Data for South Asian countries was filtered, and death rates from 2018 to 2021 were plotted for each country using line plots. This detailed examination helped identify the countries most impacted by air pollution in terms of mortality.

### 3.5. Correlation Analysis

In this subsection, we performed three key correlation analyses to explore the relationships between PM 2.5 levels, population density, and death rates across various regions in Asia. These analyses aimed to uncover significant patterns and provide deeper insights into how air pollution impacts different areas.

1) Correlation Between PM 2.5 Levels and Population Density: We calculated for the year 2023 to assess the impact of population density on PM 2.5 levels. This analysis was conducted separately for each region to determine if more densely populated areas have higher particulate matter concentrations. The expected result is to reveal whether a significant positive relationship exists, indicating that higher human activity areas may face more severe air pollution.
2) Correlation Between Average PM 2.5 Levels and Average Death Rates: To investigate the relationship between air pollution and mortality, we conducted a correlation analysis using average PM 2.5 levels and death rates. The analysis involved comparing average PM 2.5 levels from 2018 to 2023 with average death rates from 2018 to 2021 to assess their relationship. Additionally, we examined the correlation between PM 2.5 levels from 2018 to 2021 and death rates for the same period to determine if recent pollution data (2022-2023) had an impact on the results.

These correlation analyses are essential in understanding the broader impacts of PM 2.5 on human health and environmental sustainability across Asia.

### 3.6. Clustering

In this study, K-Means Clustering was used to categorize countries based on their PM 2.5 levels in 2023. The goal was to group countries with similar air quality profiles into distinct clusters for analyzing regional pollution patterns. PM 2.5 data was standardized to ensure variables were on a comparable scale before applying the clustering algorithm. We determined using the Elbow Method, which involved plotting the within-cluster sum of squares (WCSS) against various k values to identify the optimal k at the "elbow" point where additional clusters offer minimal improvement. Then we applied K-Means to the standardized data to create clusters representing different levels of air pollution. Each cluster was analyzed to understand the distribution of PM 2.5 levels across countries. This clustering process played a crucial role in identifying and understanding the patterns of air quality in the region.

### 3.7. ARIMA Model for Prediction

The ARIMA model was used to forecast PM 2.5 levels. Data from 2018 to 2022 trained the model, with its performance assessed against actual 2023 values using MAE, MSE, RMSE, and $R^2$. For 2024, predictions were made using data up to 2023; however, without actual 2024 data, evaluation metrics for this forecast are not available.

### 3.8. Evaluation Matrices

The ARIMA model's performance for 2023 was evaluated using four key metrics: Mean Absolute Error (MAE), which measures the average magnitude of prediction errors; Mean Squared Error (MSE), indicating the average squared difference between predicted and actual values; Root Mean Squared Error (RMSE), providing the square root of the average squared errors to show prediction error size; and R-squared ($R^2$), representing the proportion of variance in the dependent variable explained by the model.

## 4. RESULT ANALYSIS

### 4.1. Temporal Analysis of PM 2.5 Levels

#### 4.1.1. South Asia Trends

The temporal trends of PM 2.5 levels across South Asia from 2018 to 2023 are depicted in Figure 1. The data reveals varying air quality trends among the countries in the region.

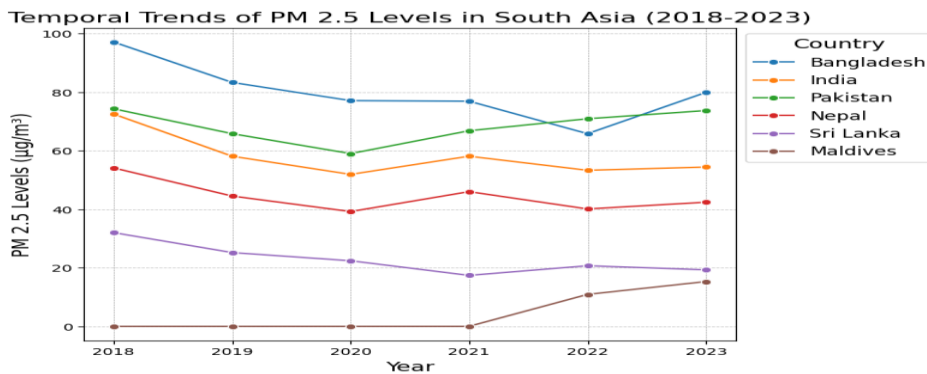

Figure 1. Temporal Trends of PM 2.5 Levels in South Asia (2018 - 2023)

Bangladesh has the highest PM 2.5 levels in South Asia, increasing from 79.9 in 2018 to 97.1 in 2023, reflecting a significant rise in pollution. India also shows an upward trend, from 54.4 to 72.5 over the same period, indicating worsening air quality. Pakistan's PM 2.5 levels, while relatively high, show a slight decrease overall, particularly from 66.8 in 2020 to 59 in 2021. Nepal's levels increase gradually from 42.4 to 54.1, while Sri Lanka maintains comparatively lower levels, rising from 19.3 to 32, both showing moderate trends. The Maldives' data is limited to 2018 (15.3) and 2019 (10.9), preventing a full assessment of recent trends. This data gap highlights the need for comprehensive monitoring, and future studies should consider methods like imputation to estimate missing values. However, it is important to acknowledge that imputation may not fully capture regional variations and could impact the accuracy of the overall trend analysis. These trends indicate varying air quality across the region, with some countries facing more severe pollution challenges than others, highlighting the need for comprehensive monitoring, especially in data-scarce areas like the Maldives.

### 4.1.2. Central Asia Trends

The temporal trends of PM 2.5 levels across Central Asia from 2018 to 2023 are illustrated in Figure 2. The data reveals a diverse range of air quality trends among the countries in this region.

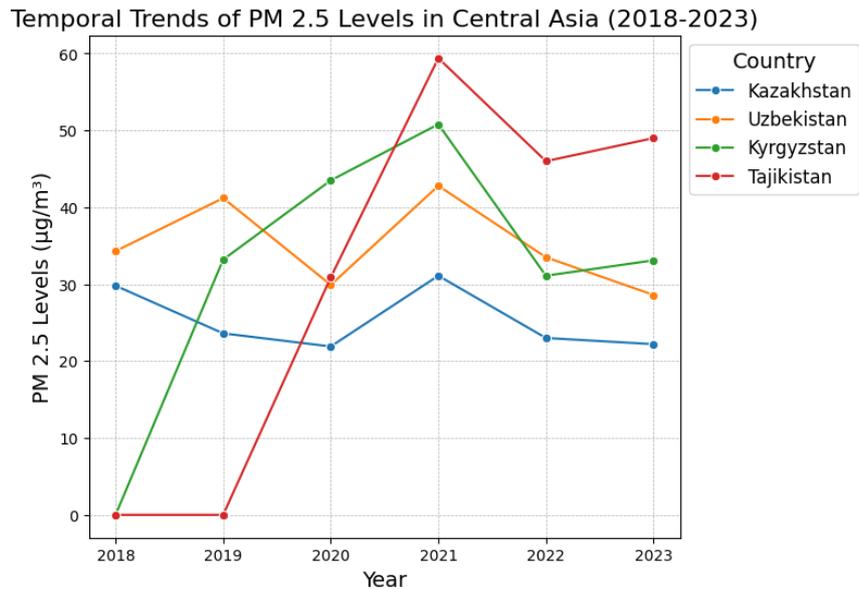

Figure 2. Temporal Trends of PM 2.5 Levels in Central Asia (2018 - 2023)

Kazakhstan shows relatively stable PM 2.5 levels, with a slight increase from 21.9 in 2021 to 29.8 in 2023. Uzbekistan displays a more variable trend, peaking at 42.8 in 2020 before declining to 34.3 in 2023. Kyrgyzstan shows a peak at 50.8 in 2020 but lacks data for subsequent years, which limits the ability to assess more recent trends. Similarly, Tajikistan's data is only available up to 2020, peaking at 59.4, preventing a comprehensive analysis of air quality in later years. The lack of data in both countries highlights the importance of improving monitoring systems in Central Asia. To address these data gaps, future studies could consider methods like imputation, though it is essential to acknowledge that such techniques may not fully capture regional variations and could affect the accuracy of the trend analysis. These trends indicate variability in air quality across Central Asia, with some countries showing significant fluctuations and others more stable levels.

**4.1.3. Southeast Asia Trends**

The temporal trends of PM 2.5 levels across Southeast Asia from 2018 to 2023 are depicted in Figure 3. The data shows considerable variability in air quality trends among the countries in this region. Indonesia shows a notable rise in PM 2.5 levels, peaking at 51.7 in 2021 before decreasing to 42 in 2023, indicating significant fluctuations in air pollution. Thailand's levels are relatively stable, rising slightly from 23.3 in 2018 to 26.4 in 2023, suggesting a slow increase in pollution. Vietnam's levels peaked at 34.1 in 2021 before dropping to 32.9 in 2023, reflecting moderate fluctuations. The Philippines maintains low levels, ranging from 13.5 in 2018 to 14.6 in 2023, showing stable air quality.

Malaysia's data indicates a decline from 22.5 in 2018 to 19.4 in 2021, with missing values for 2023. Singapore maintains consistently low levels, fluctuating slightly from 13.4 in 2018 to 14.8 in 2023. Myanmar and Laos lack recent data, but previous levels show Myanmar rising slightly and Laos decreasing. Cambodia's levels dropped sharply from 22.8 in 2018 to 8.3 in 2019, then rose again to 21.1 in 2021 before slightly decreasing to 20.1 in 2023, showing high variability. Overall, Southeast Asia displays diverse air quality trends, with some countries facing increasing pollution and others showing stability or decline, highlighting the need for tailored regional policies.

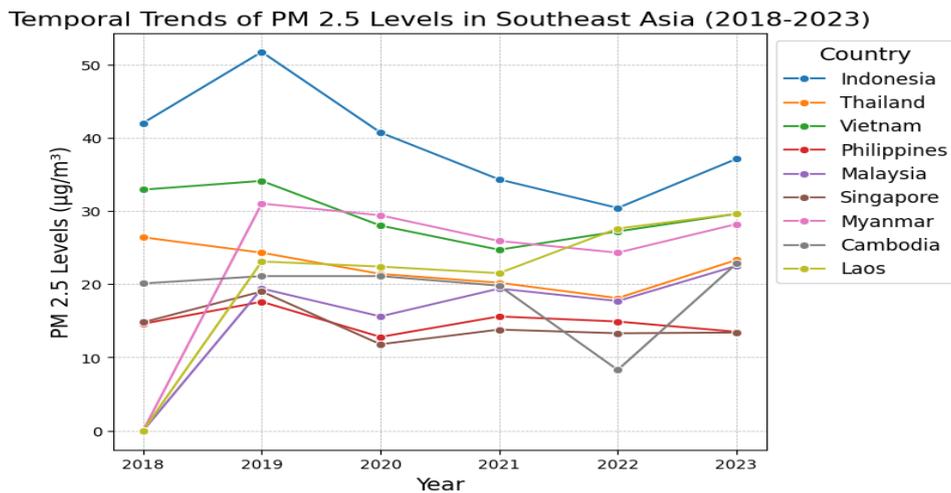

Figure 3. Temporal Trends of PM 2.5 Levels in Southeast Asia (2018 - 2023)

**4.1.4. East Asia Trends**

The temporal trends of PM 2.5 levels across East Asia from 2018 to 2023 are presented in Figure 4. The data reveals varied trends in air quality across the countries in this region.

China shows a rising trend in PM 2.5 levels, increasing from 32.5 in 2018 to 42.2 in 2023, reflecting ongoing pollution challenges. Japan maintains the lowest levels in East Asia, ranging from 9.1 in 2018 to 12 in 2023, indicating stable air quality. South Korea's levels gradually rise from 19.2 in 2018 to 24 in 2023, pointing to growing air quality concerns. Taiwan's PM 2.5 levels fluctuate, dropping from 20.2 in 2018 to 15 in 2020, then rising to 18.5 in 2023, suggesting variability in pollution trends. Mongolia records the highest levels, sharply rising from 22.5 in 2018 to 62 in 2021, and slightly dropping to 58.5 in 2023, indicating severe air quality issues. Macao SAR sees a moderate increase from 16.2 in 2018 to 21.2 in 2023, and Hong Kong SAR's levels also rise from 14.5 in 2018 to 20.2 in 2023, both reflecting worsening air quality. These trends show diverse pollution levels across East Asia, emphasizing the need for targeted, region-specific interventions to improve air quality.

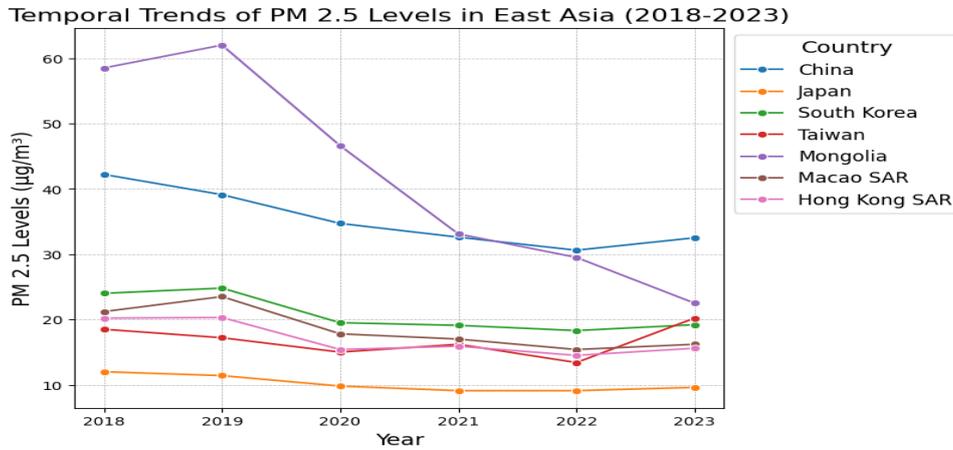

Figure 4. Temporal Trends of PM 2.5 Levels in East Asia (2018 - 2023)

### 4.1.5. West Asia Trends

The temporal trends of PM 2.5 levels across West Asia from 2018 to 2023 are illustrated in Figure 5. The data highlights a wide range of air quality trends among the countries in this region. Turkey maintains relatively stable PM 2.5 levels with slight increases from 18.7 in 2021 to 21.9 in 2023. Saudi Arabia shows significant fluctuation, peaking at 41.5 in 2019 and dropping to 22.1 in 2021, with missing data for 2023. Iraq's levels vary dramatically, peaking at 80.1 in 2019 and falling to 39.6 in 2021, also with incomplete data.

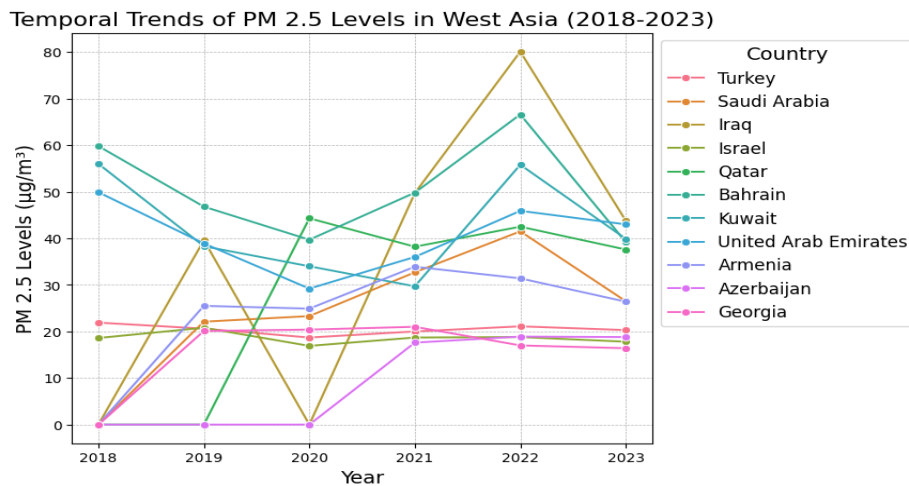

Figure 5. Temporal Trends of PM 2.5 Levels in West Asia (2018 - 2023)

Israel's levels remain relatively stable, ranging from 16.9 in 2021 to 18.6 in 2023. Qatar's incomplete data shows an increase from 37.6 in 2018 to 44.3 in 2021. Bahrain experiences a sharp rise from 39.2 in 2018 to 59.8 in 2023, indicating worsening air quality. Kuwait fluctuates, peaking at 55.8 in 2019, dropping to 34 in 2021, and rising again to 56 in 2023. The UAE's levels decline from 43 in 2018 to 29.2 in 2021 but increase to 49.9 in 2023. Armenia shows an upward trend, from 26.4 in 2018 to 33.9 in 2020, with gaps in later data. Azerbaijan and Georgia also have incomplete data but show moderate PM 2.5 levels. These trends highlight diverse air quality conditions in West Asia, with significant variability and incomplete monitoring in several countries. The incomplete data for several countries emphasizes the need for more detailed monitoring to fully understand regional air quality dynamics.

### 4.1.6. Overall Asia Trends

Figure 6 illustrates the temporal trends of average PM 2.5 levels across different regions in Asia from 2018 to 2023. The data showcases varying patterns of PM 2.5 levels, reflecting the diverse environmental conditions and pollution control efforts within the continent.

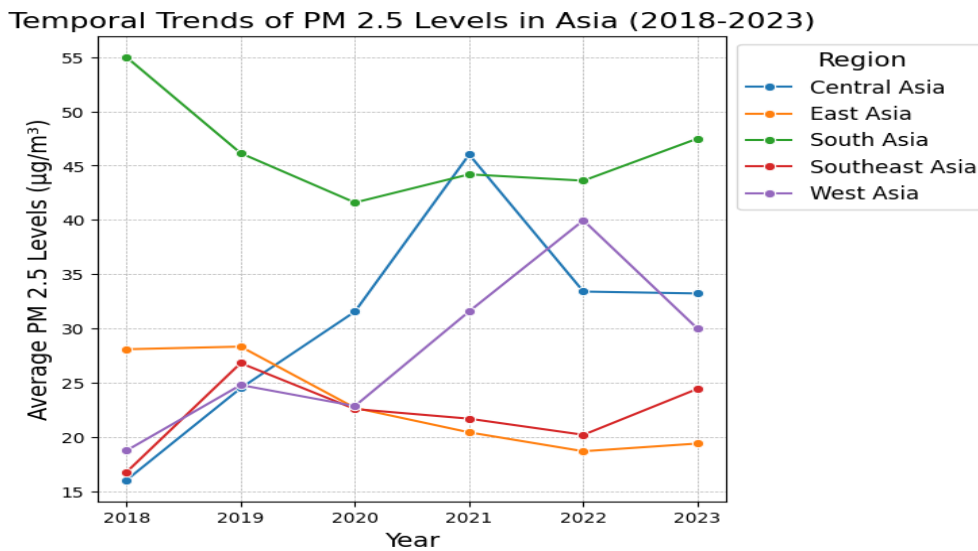

Figure 6. Temporal Trends of PM 2.5 Levels in Overall Asia (2018 - 2023)

South Asia has the highest PM 2.5 levels among all regions, showing a slight decline from 2018 to 2020, but rising again from 2022 to 2023. Central Asia experiences fluctuations, peaking in 2021, then declining as pollution control measures likely took effect. Southeast Asia's PM 2.5 levels steadily increase until peaking in 2021, followed by a decline, suggesting improved air quality due to stricter regulations. East Asia and West Asia maintain the lowest PM 2.5 levels, with stable trends; however, East Asia shows a slight rise post-2021, while West Asia remains steady after a minor increase in 2019. The diverse trends observed across the different regions of Asia highlight the varying impact of air pollution across the continent. However, the gaps in data for countries such as the Maldives, Kyrgyzstan, and more limit the ability to draw comprehensive conclusions for the entire region. Future research should aim to address these data gaps, either through improved monitoring or statistical methods like imputation, to ensure more robust assessments and informed policymaking. The overall trends suggest that while some regions have made significant strides in improving air quality, others continue to face challenges, necessitating ongoing and targeted efforts to reduce PM 2.5 levels and protect public health.

## 4.2. Death Rate Analysis

### 4.2.1. Regional Death Rate Trends (2018-2021)

From 2018 to 2021, death rates due to air pollution varied significantly across Asia as shown in Figure 7. South Asia consistently recorded the highest death rates, reflecting severe pollution impacts in the region. Central Asia also exhibited high death rates, while Southeast and West Asia maintained moderate levels with some fluctuations. East Asia consistently had the lowest death rates, suggesting effective pollution control measures. In 2019, South Asia continued to have the highest rates, while Central Asia saw a slight reduction. A general decline in death rates was observed in 2020 across most regions, but rates increased again in 2021, particularly in Southeast and Central Asia. Overall, these trends underscore the persistent challenge of managing air pollution and its health impacts, especially in South Asia.

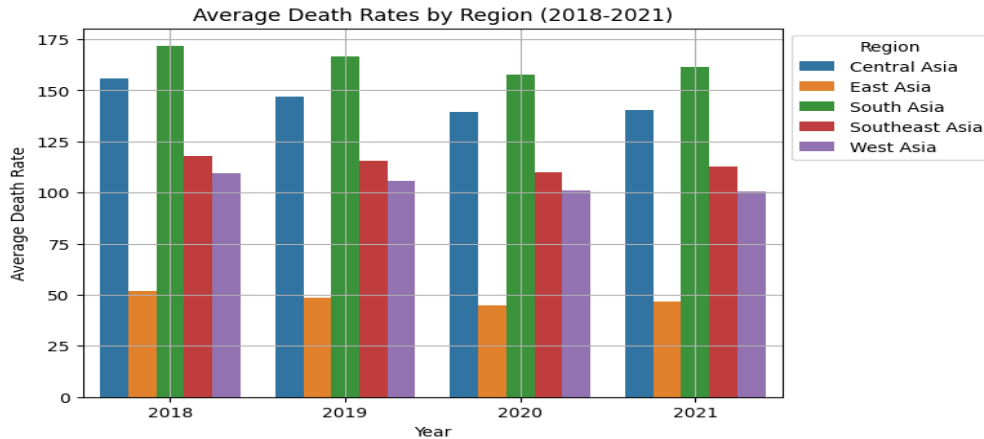

Figure 7. Death Rate of Overall Asia (2018 - 2021)

### 4.2.2. Death Rate Trends in South Asia (2018-2021)

The line plot in Figure 8 for death rates in South Asian countries from 2018 to 2021 shows that Nepal consistently has the highest death rates throughout the period. Pakistan follows, with India and Bangladesh showing slightly lower but still relatively high death rates. Sri Lanka exhibits a moderate trend, while Maldives consistently has the lowest death rates among the countries. The trends indicate that the countries with higher death rates have remained fairly stable, with only slight fluctuations over the years.

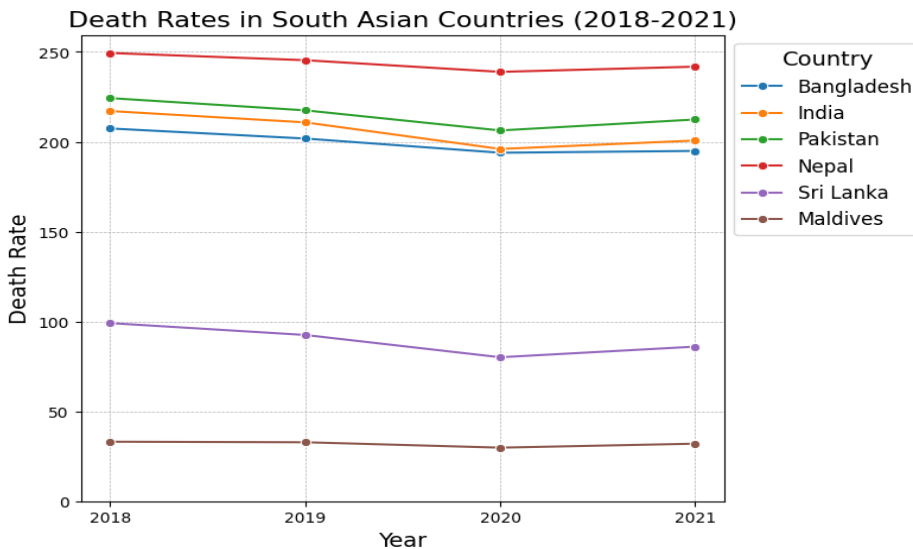

Figure 8. Death Rate of South Asia (2018 - 2021)

International Journal on Cybernetics & Informatics (IJCI) Vol.14, No.1, February 2025

## 4.3. Correlation Analysis

In this section, we explored the relationships between PM 2.5 levels, population density, and death rates across various regions in Asia. By examining these correlations, we aim to uncover significant patterns and insights into how air pollution impacts different regions and populations.

### 4.3.1. Correlation Between PM 2.5 Levels and Population Density

This analysis examined the relationship between population density and PM 2.5 levels for 2023 across different regions to see if higher population densities correlate with increased air pollution. The correlation coefficient was approximately -0.20, indicating a weak negative relationship. This suggests that regions with higher population densities do not necessarily have higher PM 2.5 levels. The heatmap visualization shown in Figure 9 confirmed that other factors, such as geography, industrial activities, and local environmental policies, might have a more substantial impact on air pollution than population density alone. These findings imply that effective policy interventions should consider multiple factors beyond population density to address air quality issues.

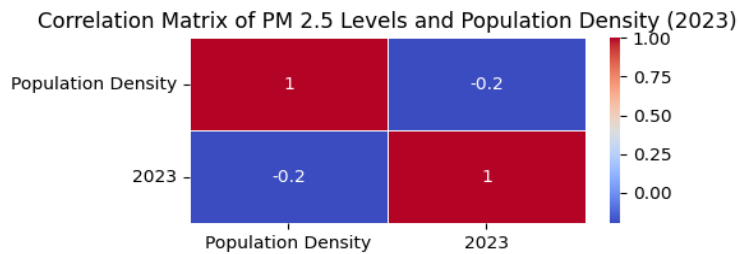

Figure 9. Correlation Matrix of PM 2.5 Levels and Population Density

### 4.3.2. Correlation Between Average PM 2.5 Levels and Average Death Rates

The correlation between average PM 2.5 levels from 2018 to 2023 and death rates from 2018 to 2021 is 0.63, as shown in Figure 10, indicating a moderate positive link between higher pollution and higher mortality. In comparison, the correlation of 0.57, was based on data from 2018 to 2021 alone. The stronger correlation with the more recent data suggests a growing impact of sustained high pollution levels on health, emphasizing the importance of including the latest data for a more accurate assessment.

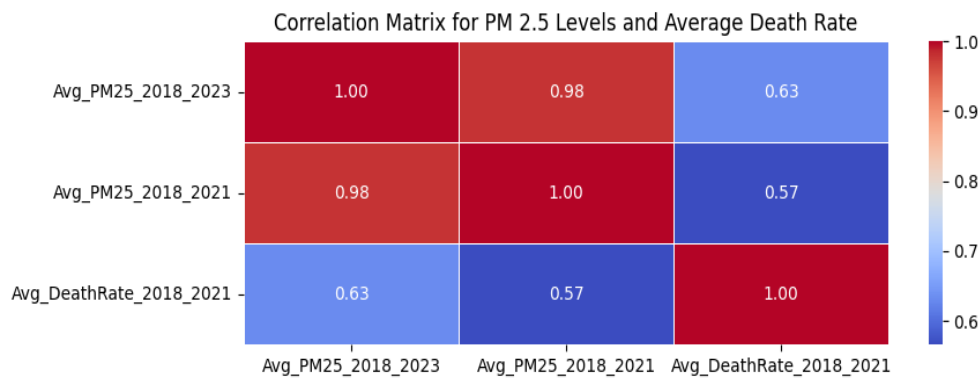

Figure 10. Correlation Matrix of Average PM 2.5 Levels (2018 - 2023 and 2018 - 2021) and Average Death Rate (2018 - 2021)

International Journal on Cybernetics & Informatics (IJCI) Vol.14, No.1, February 2025

## 4.4. Clustering Analysis of PM 2.5 Levels

The clustering analysis aimed to categorize Asian countries based on their PM 2.5 levels in 2023 using the K-Means algorithm. The primary objective was to identify distinct clusters that represent different levels of air pollution, categorized as low, moderate, and high pollution. The Elbow Method, illustrated in Figure 11, identified three as the optimal number of clusters. Each cluster was labelled according to the PM 2.5 levels: "Low Pollution," "Moderate Pollution," and "High Pollution."

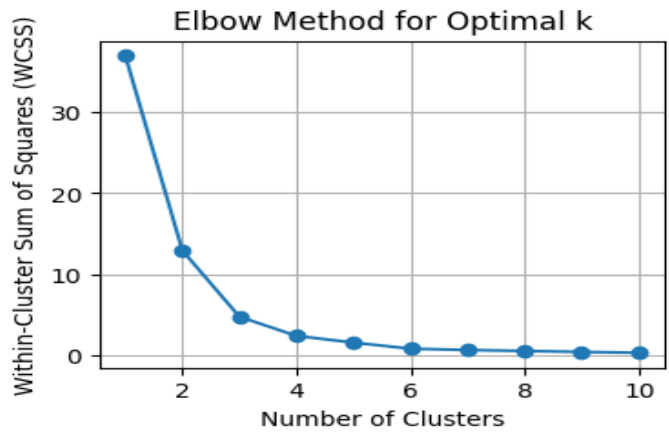

Figure 11. Elbow Method for Optimal k

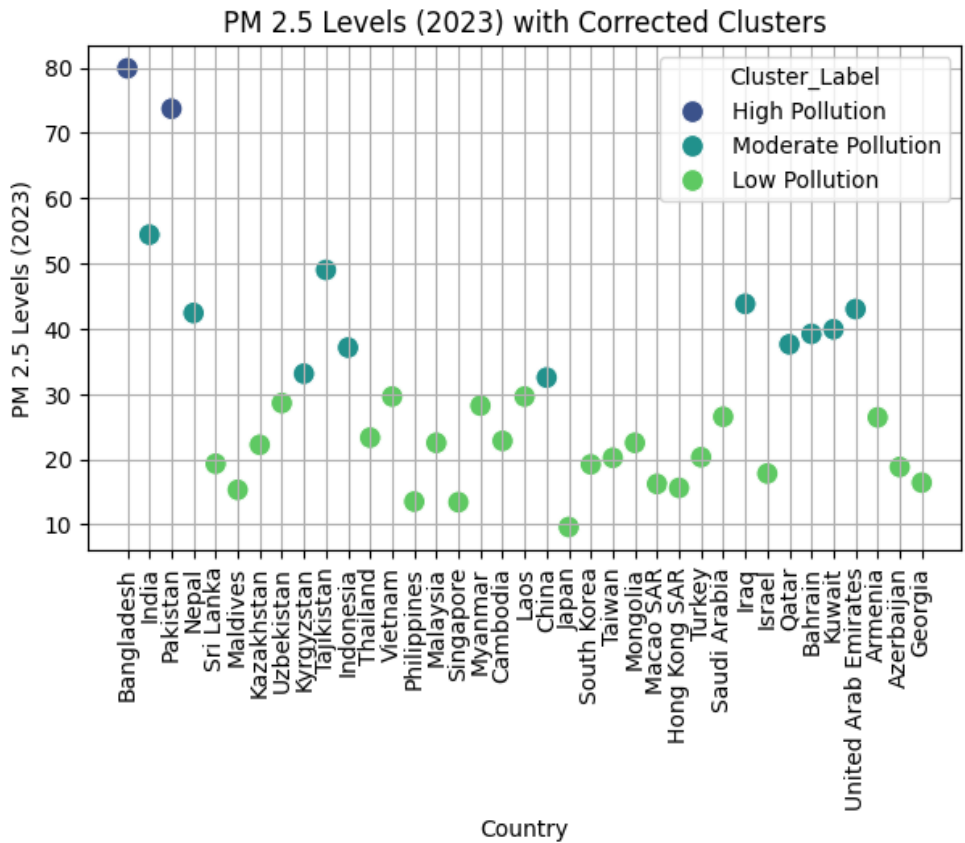

Figure 12. PM 2.5 Levels with Corrected Clusters

The results of the clustering are visualized in Figure 12, where countries are plotted against their PM 2.5 levels in 2023, color-coded by their respective cluster labels. This visualization effectively differentiates the countries based on their pollution levels, with each cluster representing a distinct range of PM 2.5 levels. Table I summarizes the PM 2.5 level ranges corresponding to each cluster label.

Table 1.  Cluster centres based on pm 2.5 levels (2023)

| Cluster levels | PM 2.5 levels (2023) |
|---|---|
| Low pollution | 20.74 |
| Moderate pollution | 41.09 |
| High pollution | 76.80 |

This clustering analysis not only provides a clear categorization of countries based on their PM 2.5 pollution levels but also offers a framework for targeted policy interventions. For countries in the "High Pollution" cluster, immediate actions such as stricter emission controls, public health advisories, and investment in air purification systems may be necessary. For the "Moderate Pollution" cluster, governments could focus on implementing regulatory measures to curb pollution sources, while also monitoring the air quality closely. Finally, countries in the "Low Pollution" cluster might prioritize sustaining their current pollution levels by reinforcing policies that promote cleaner technologies and renewable energy sources. The clustering results, therefore, offer valuable insights for policymakers, enabling them to design region-specific and pollution level-tailored strategies to improve air quality across Asia.

### 4.5. Prediction (ARIMA) and Evaluation Metrics

1) ARIMA Model Training and Forecasting Process: The ARIMA (AutoRegressive Integrated Moving Average) model was utilized to forecast PM 2.5 levels for 2023 using 2018- 2022 data for training, with predictions compared to actual 2023 values to assess accuracy. For 2024 predictions, the model was trained on 2018-2023 data, but these forecasts await validation as actual 2024 data is unavailable.
2) Evaluation Metrics for 2023: The performance of the ARIMA model was assessed using the following metrics:

Table 2.  Evaluation metrics for the ARIMA model in 2023

| Metric | Value |
|---|---|
| Mean Absolute Error (MAE) | 3.99 |
| Mean Squared Error (MSE) | 33.80 |
| Root Mean Squared Error (RMSE) | 5.81 |
| R-squared (R²) | 0.86 |

Table 2 shows ARIMA model's performance for 2023 was evaluated using several metrics. The Mean Absolute Error (MAE) of 3.99 units, Mean Squared Error (MSE) of 33.80, and Root Mean Squared Error (RMSE) of 5.81 suggest that the model's predictions were generally accurate, with lower values indicating fewer and smaller errors. The R² value of 86% shows that the model effectively explained the variance in PM 2.5 levels for 2023. These metrics indicate accurate predictions and strong performance in capturing PM 2.5 variance. Further validation with 2024 data is required to assess long-term accuracy.

# 5. CONCLUSION

This paper analyzes PM 2.5 levels across Asia from 2019 to 2023 using the ARIMA model, highlighting trends and regional disparities among 36 countries. Countries were classified into low, moderate, and high pollution levels for 2023, providing insights to guide policy development. While this study focuses on Asia, future research will extend globally to include regions like Europe and Africa. Additionally, integrating other predictive algorithms is planned to enhance analysis robustness and provide a more comprehensive comparison. These findings underscore the importance of continuous air quality monitoring and predictive modeling to assess health impacts and inform mitigation strategies. Ultimately, the study advocates for collaborative, data-driven approaches among policymakers, environmental agencies, and researchers to address air pollution in rapidly urbanizing regions.

**AUTHORS**

*Anika Rahman:* Anika Rahman completed her bachelor's degree from Ahasanullah University of Science and Technology, Bangladesh in the Department of Computer Science and Engineering and her Master's degree from BRAC University, Bangladesh. Currently, she is working as a Lecturer in the Department of Computer Science and Engineering at Stamford University Bangladesh. Her research area focuses on AI, machine learning, and Data Science. 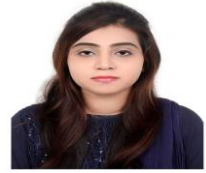

*Dr. Mst. Taskia Khatun:* Dr. Mst. Taskia Khatun completed her bachelor's degree in Computer Science and Engineering at Rajshahi University of Engineering & Technology, Bangladesh. After That, she completed her Master's and PhD from The University of Tokyo, Japan. Currently, she is working as an Assistant Professor in the Department of Software Engineering at Daffodil International University, Bangladesh. Her research area focuses on project management, system analysis and design, simulation and modelling, data science. 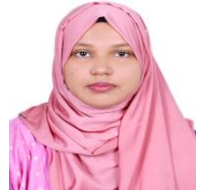